# Diffusion Maximum Correntropy Criterion Algorithms for Robust Distributed Estimation

**Wentao Ma[a], Badong Chen[b],*, Jiandong Duan[a], Haiquan Zhao[c]**

[a]Department of Electrical Engineering, Xi'an University of Technology, Xi'an 710048, China
[b]School of Electronic and Information Engineering, Xi'an Jiaotong University, Xi'an, 710049, China
[c] School of Electrical Engineering, Southwest Jiaotong University, Chengdu, China

*Correspondence author：Badong Chen. E-Mail: chenbd@mail.xjtu.edu.cn

**Abstract:**

Robust diffusion adaptive estimation algorithms based on the maximum correntropy criterion (MCC), including adaptation to combination MCC and combination to adaptation MCC, are developed to deal with the distributed estimation over network in impulsive (long-tailed) noise environments. The cost functions used in distributed estimation are in general based on the mean square error (MSE) criterion, which is desirable when the measurement noise is Gaussian. In non-Gaussian situations, such as the impulsive-noise case, MCC based methods may achieve much better performance than the MSE methods as they take into account higher order statistics of error distribution. The proposed methods can also outperform the robust diffusion least mean p-power(DLMP) and diffusion minimum error entropy (DMEE) algorithms. The mean and mean square convergence analysis of the new algorithms are also carried out.

**Keywords:**

Correntropy; maximum correntropy criterion; diffusion; robust distributed adaptive estimation

## 1. Introduction

As an important issue in the field of distributed network, the distributed estimation over network plays a key role in many applications, including environment monitoring, disaster relief management, source localization, and so on [1-4], which aims to estimate some parameters of interest from noisy measurements through cooperation between nodes. Much progress has been made in the past few years.

In particular, the diffusion mode of cooperation for distributed network estimation(DNE) has aroused more and more concern among researchers, which keeps the nodes exchange their estimates with neighbors and fuses the collected estimates via linear combination. So far a number of diffusion mode algorithms have been developed by researchers, such as the diffusion least mean square (DLMS) [5-8], diffusion recursive least square (DRLS)[9] and their variants [10-13]. These algorithms are derived under the popular mean square error (MSE) criterion, of which the optimizations are well understood and efficient. It is well-known that the optimality of MSE relies heavily on the Gaussian and linear assumptions. In practice, however, the data distributions are usually non-Gaussian, and in these situations, the MSE is possibly no longer an appropriate one especially in the presence of heavy-tailed non-Gaussian noise [14]. In distributed networks, some impulsive noises are usually unavoidable.

Recently, some researchers focus on improving robustness of DNE methods. The efforts are mainly directed at searching for a more robust cost function to replace the MSE cost (which is sensitive to large outliers due to the square operator). To address this problem, the diffusion least mean p-power (DLMP) based on p-norm error criterion was proposed to estimate the parameters of the wireless sensor networks [15]. For non-Gaussian cases, *Information Theoretic Learning* (ITL) [16] provides a more general framework and can also achieve desirable performance. The diffusion minimum error entropy (DMEE) was proposed in [17]. Under the MEE criterion, the entropy of a batch of N most recent error samples is used as a cost function to be minimized to adapt the weights. The evaluation of the error entropy involves a double sum over the samples, which is computationally expensive especially when the window length $L$ is large.

In recent years, the correntropy as a nonlinear similarity measure in ITL, has been successfully used as a robust and efficient cost function for non-Gaussian signal processing [18]. The adaptive algorithms under the maximum correntropy criterion (MCC) are shown to be very robust with respect to impulsive noises, since correntropy is a measure of local similarity and is insensitive to outliers [19]. Moreover, MCC based algorithms are, in general, computationally much simpler than the MEE based algorithms. Research results on dimensionality reduction[20], feature selection [21], robust regression [22] and adaptive filtering [23-28] have demonstrated the effectiveness of MCC when dealing with occlusion and corruption problems.

Motivated by the desirable features of correntropy, we propose in this work a novel diffusion scheme, called *diffusion MCC* (DMCC), for robust distributed estimation in impulsive noise environments. The main contributions of the paper are three-folds: (i) a correntropy-based diffusion scheme is proposed to solve the distributed estimation over networks; (ii) two MCC based diffusion algorithms, namely *adaptation to combination* (ATC) and *combination to adaptation* (CTA) diffusion algorithms are developed, which can combat impulsive noises effectively; (iii) the mean and mean square performances have been analyzed. Moreover, simulations are conducted to illustrate the performance of the proposed methods under impulsive noise disturbances.

The remainder of the paper is organized as follows. In Section 2, we give a brief review of MCC. In Section 3, we propose the DMCC method and present two adaptive combination versions. The mean and mean square analysis are performed in section 4. Simulation results are then presented in section 5 Finally, conclusion is given in Section 6.

## 2. Maximum correntropy criterion

The correntropy between two random variables $x$ and $y$ is defined by

$$V(x, y) = E[\kappa(x, y)] = \int \kappa(x, y) dF_{xy}(x, y) \quad (1)$$

where E[.] denotes the expectation operator, $\kappa(\cdot,\cdot)$ is a shift-invariant Mercer kernel, and $F_{xy}(x, y)$ denotes the joint distribution function. In practice, only a finite number of samples $\{x_i, y_i\}_{i=1}^{N}$ are available, and the joint distribution is usually unknown. In this case, the correntropy can be estimated as the sample mean

$$V(x, y) = E[\kappa(x, y)] \approx \frac{1}{N} \sum_{i=1}^{N} \kappa(x_i, y_i) \quad (2)$$

The most popular kernel used in correntropy is the Gaussian kernel:

$$\kappa_\sigma(x, y) = \frac{1}{\sigma\sqrt{2\pi}} \exp(-\frac{e^2}{2\sigma^2}) \quad (3)$$

where $e = x - y$, and $\sigma$ denotes the kernel size. With Gaussian kernel, the instantaneous MCC cost is [18]

$$J_{MCC}(i) = G_\sigma^{MCC}(e(i)) = \frac{1}{\sigma\sqrt{2\pi}} \exp(-\frac{e^2(i)}{2\sigma^2}) \quad (4)$$

where $i$ denotes the time instant (or iteration number). MCC (with Gaussian kernel) has some desirable properties[19]: 1) it is always bounded for any distribution; 2) it contains all even-order moments, and the weights of the higher-order moments are determined by the kernel size; 3) it is a local similarity measure and is robust to outliers. Based on these excellent properties, we develop the diffusion MCC algorithms in the next section.

### 3. Diffusion MCC algorithms

*3.1. General diffusion MCC*

Consider a network composed of $N$ nodes distributed over a geographic area to estimate an unknown vector $w_o$ of size($M \times 1$) from measurements collected at $N$ nodes. At each time instant $i$ ($i = 1, 2, \cdots I$), each node $k$ has access to the realization of a scalar measurement $d_k$ and a regression vector $u_k$ of size($M \times 1$), related as

$$d_k(i) = w_o^T u_k(i) + n_k(i) \quad (5)$$

where $n_k(i)$ denotes the measurement noise, and $T$ denotes transposition.

Given the above model, for each node $k$, the DMCC seeks to estimate $w_o$ by maximizing a linear combination of the local correntropy within the node $k$'s neighbor $N_k$. The cost function of the DMCC for each node can be therefore expressed as

$$J_k^{local}(w) = \sum_{l \in N_k} \alpha_{l,k} G_\sigma^{MCC}(e_{l,k}(i))$$
$$= \sum_{l \in N_k} \alpha_{l,k} G_\sigma^{MCC}(d_l(i) - w^T u_l(i)) \quad (6)$$

where $w$ is the estimate of $w_o$, $e_{l,k}(i) = d_l(i) - w^T u_l(i)$, $\{\alpha_{l,k}\}$ are some non-negative combination coefficients satisfying $\sum_{l \in N_k} \alpha_{l,k} = 1$, and $\alpha_{l,k} = 0$ if $l \notin N_k$, and

$$G_\sigma^{MCC}(e_{l,k}(i)) = \frac{1}{\sigma\sqrt{2\pi}} \exp(-\frac{1}{2\sigma^2}(e_{l,k}(i))^2)$$
$$= \frac{1}{\sigma\sqrt{2\pi}} \exp(-\frac{1}{2\sigma^2}(d_l(i) - w^T u_l(i))^2) \quad (7)$$

Taking the derivative of (6) yields

$$\nabla J_k^{local}(w) = \sum_{l \in N_k} \alpha_{l,k} \frac{\partial G_\sigma^{MCC}(e_{l,k}(i))}{\partial w}$$
$$= \frac{1}{\sigma^2} \sum_{l \in N_k} \alpha_{l,k} G_\sigma^{MCC}(e_{l,k}(i)) e_{l,k}(i) u_l(i) \quad (8)$$

A gradient based algorithm for estimating $w_o$ at node $k$ can thus be derived as

$$w_k(i) = w_k(i-1) + \mu_k \nabla J_k^{local}(w)$$
$$= w_k(i-1) + \frac{\mu_k}{\sigma^2} \sum_{l \in N_k} \alpha_{l,k} G_\sigma^{MCC}(e_{l,k}(i)) e_{l,k}(i) u_l(i) \quad (9)$$

where $w_k(i)$ stands for the estimate of $w_o$ at time instant $i$, and $\mu_k$ is the step size for node $k$. There are mainly two different schemes (including the adapt-then-combine (ATC) scheme and the combine-then-adapt (CTA) scheme) for the diffusion estimation in the literature[6,8]. The ATC scheme first updates the local estimates using the adaptive algorithm and then the estimates of the neighbors are fused together, while the CTA scheme [7] performs the operations of the ATC scheme in a reverse order. In the next 4.3 section, we will give these two version of DMCC algorithms. For each node, we calculate the intermediate estimates by

$$\varphi_k(i-1) = \sum_{l \in N_k} \beta_{l,k} w_l(i-1) \quad (10)$$

where $\varphi_k(i-1)$ denotes an intermediate estimate offered by node $k$ at instant $i-1$, and $\beta_{l,k}$ denotes a weight with which a node should share its intermediate estimate $w_l(i-1)$ with node $k$. With all the intermediate estimates, the nodes update their estimates by

$$\phi_k(i) = \varphi_k(i-1) + \frac{\mu_k}{\sigma^2} \sum_{l \in N_k} \alpha_{l,k} G_\sigma^{MCC}(e_{l,k}) e_{l,k} u_l(i) \quad (11)$$

Above iteration in (11) is referenced as incremental step. The coefficients $\{\alpha_{l,k}\}$ determine which nodes should share their measurements $\{d_l(i), u_l(i)\}$ with node $k$.

The combination is then performed as

$$w_k(i) = \sum_{l \in N_k} \delta_{l,k} \phi_l(i) \quad (12)$$

This result in (12) represents a convex combination of estimates from incremental step (11) fed by spatially distinct data $\{d_k(i), u_k(i)\}$, and it is referenced as diffusion step. The coefficients in $\{\delta_{l,k}\}$ determine which nodes should share their intermediate estimates $\phi_l(i)$ with node $k$.

According to above analysis, one can obtain the following general diffusion MCC method by combining (9),(10) and (11)

$$\begin{cases} \varphi_k(i-1) = \sum_{l \in N_k} \beta_{l,k} w_l(i-1) & diffusion \ I \\ \phi_k(i) = \varphi_k(i-1) + \eta_k \sum_{l \in N_k} \begin{bmatrix} \alpha_{l,k} G_\sigma^{MCC}(d_l(i) - u_l(i)\varphi_k(i-1)) \\ (d_l(i) - u_l(i)\varphi_k(i-1))u_l(i) \end{bmatrix} & incremental \\ w_k(i) = \sum_{l \in N_k} \delta_{l,k} \phi_l(i) & diffusion \ II \end{cases} \quad (13)$$

where $\eta_k = \frac{\mu_k}{\sigma^2} (k=1,2,\cdots N)$. Details on the selection of the weights $\beta_{l,k}$, $\alpha_{l,k}$, and $\delta_{l,k}$ can be found in [8].

**Remark1:** One can see that the equation (13) contains an extra scaling factor $G_\sigma^{MCC}(e_{l,k}(i))$, which is an exponential function of the error. When a large error occurs (possibly caused by an outlier), this scaling factor will approach zero, which endows the DMCC with the outlier rejection property and will improve significantly the adaptation performance in impulsive noises.

**Remark2:** The kernel size $\sigma$ has significant influence on the performance of the DMCC, similar to most kernel methods. In general, a larger kernel size makes the algorithm less robust to the outliers, while a smaller kernel size makes the algorithm stall.

3.2 *ATC and CTA diffusion MCC*

The non-negative real coefficients $\{\beta_{l,k}\}$, $\{\alpha_{l,k}\}$, $\{\delta_{l,k}\}$ in (13) are corresponding to the $\{l,k\}$ entries of matrices $\mathbf{P}_1$, $\mathbf{P}_2$ and $\mathbf{P}_3$, respectively, and satisfy

$$\mathbf{1}^T \mathbf{P}_1 = \mathbf{1}^T, \mathbf{1}^T \mathbf{P}_2 = \mathbf{1}^T, \mathbf{1}^T \mathbf{P}_3 = \mathbf{1}^T$$

where $\mathbf{1}$ denotes the $N \times 1$ vector with unit entries. Below we develop the ATC and CTA diffusion MCC algorithms.

**ATC diffusion MCC:** When $\mathbf{P}_1 = I, \mathbf{P}_2 = I$, the algorithm (13) will reduce to the uncomplicated ATC diffusion MCC (ATCDMCC) version as

$$\begin{cases} \phi_k(i) = w_k(i-1) + \eta_k \begin{bmatrix} G_\sigma^{MCC}(d_k(i) - u_k(i)w_k(i-1)) \\ (d_k(i) - u_k(i)w_k(i-1))u_k(i) \end{bmatrix} \\ w_k(i) = \sum_{l \in N_k} \delta_{l,k} \phi_l(i) \end{cases} \quad (14)$$

**CTA diffusion MCC:** Similar to the ATC version, one can get a simple CTA diffusion MCC (CTADMCC) algorithm by choosing $\mathbf{P}_2 = I$ and $\mathbf{P}_3 = I$:

$$\begin{cases} \varphi_k(i-1) = \sum_{l \in N_k} \beta_{l,k} w_l(i-1) \\ w_k(i) = \varphi_k(i-1) + \eta_k \begin{bmatrix} G_\sigma^{MCC}(d_k(i) - u_k(i)\varphi_k(i-1)) \\ (d_k(i) - u_k(i)\varphi_k(i-1))u_k(i) \end{bmatrix} \end{cases} \quad (15)$$

The equations of (14) and (15) are similar to the ATC diffusion LMS (ATCLMS)[8], and the CTA diffusion LMS (CTALMS) [6], respectively. Clearly, the ATCDMCC and CTADMCC can be viewed as the ATCDLMS and CTADLMS with a variable step size $\frac{\mu_k}{\sigma^3 \sqrt{2\pi}} \exp(-\frac{e_k^2}{2\sigma^2})$, where $e_k$ is $d_k(i) - u_k(i)w_k(i-1)$ and $d_k(i) - u_k(i)\varphi_k(i-1)$ for ATC and CTA versions, respectively. Further, as kernel size $\sigma \to \infty$, we have $G_\sigma^{MCC}(e_{l,k}(i)) \to \frac{1}{\sigma \sqrt{2\pi}}$, which leads to the ATC and CTA diffusion LMS with fixed step size $\frac{\mu_k}{\sigma^3 \sqrt{2\pi}}$. In addition, no exchange of data is needed during the adaptation of the step size, which makes the communication cost relatively low.

**Remark3:** The ATC version usually outperforms the CTA version [7]. Similarly, the ATCDMCC algorithm tends to outperform the CTADMCC. According to (14) and (15), we know that for computing a new estimate, the ATCDMCC uses the measurement from all nodes m in the neighborhood of nodes l, which are neighbors of $k$. Thus, the ATC version effectively uses data from nodes that are two hops away in every iterations, while the CTA version uses data from nodes that are one hop away. This will be illustrated in the simulation part.

**Remark4:** The number of nodes connected to the node $k$ is denoted by $|N_k|$. The computational complexity of the ATCLMS for node $k$ at each time includes $(|N_k|+2)M+1$ multiplications and $(|N_k|+1)M$ additions [29]. For the proposed ATCDMCC, an extra computational cost is the evaluation of the exponential function of the error, which is not expensive. Thus the new methods are also computationally efficient for DNE problem.

## 4. Performance analysis

In the following, we study the convergence performance of the proposed ATCDMCC algorithm (14). The analysis of the CTADMCC algorithm is similar but not studied here. For tractable analysis, we adopt the following assumptions:

*Assumption 1:* All regressors $u_k$ arise from Gaussian sources with zero-mean and spatially and temporally independent.

**Assumption 2:** The error nonlinearity $G_\sigma^{MCC}(e_{l,k}(i))$ is independent of the regressors $u_k$.

Since nodes exchange data amongst themselves, their current update will then be affected by the weighted average of the previous estimates. Therefore, to account for this inter-node dependence, it is suitable to study the performance of the whole network. Some new variables need to be introduced. The proposed ATCDMCC algorithm can be expressed as

$$\begin{cases} \varphi_k(i) = w_k(i-1) + \eta_k \gamma_k(i) e_k(i) u_k(i) \\ \qquad = w_k(i-1) + \rho_k(i) e_k(i) u_k(i) \\ w_k(i) = \sum_{l \in N_k} \alpha_{l,k} \varphi_l(i) \end{cases} \quad (16)$$

where $\gamma_k(i) = G_\sigma^{MCC}(d_k(i) - u_k(i) w_k(i-1))$, and $\rho_k(i) = \eta_k \gamma_k(i)$ as a new step size factor. Furthermore, some other new variables need to be introduced and the local ones are transformed into global variables as follows:

$$W(i) = col\{w_1(i), w_2(i), \cdots w_N(i)\} \quad (17)$$
$$\Phi(i) = col\{\varphi_1(i), \varphi_2(i), \cdots \varphi_N(i)\} \quad (18)$$
$$U(i) = diag\{u_1(i), u_2(i), \cdots u_N(i)\} \quad (19)$$
$$\Upsilon(i) = diag\{\rho_1(i), \rho_2(i), \cdots \rho_N(i)\} \quad (20)$$
$$D(i) = col\{d_1(i), d_2(i), \cdots d_N(i)\} \quad (21)$$
$$V(i) = col\{v_1(i), v_2(i), \cdots v_N(i)\} \quad (22)$$

According the defined new variables above, a completely new set of equations representing the entire network is formed, starting with the relation between the measurements

$$D(i) = U(i) W_o + V(i) \quad (23)$$

where $W_o = I w_o$, and $I = col\{I_M, I_M, \cdots I_M\}_{MN \times M}$ is $MN \times M$ matrix. Then, the update equations can be remodeled to represent the global network

$$\begin{aligned} \Phi(i) &= W(i-1) + \Upsilon(i) U^T(i)(D(i) - U(i) W(i-1)) \\ \Upsilon(i) &= \eta_k \Omega(i) \\ W(i) &= B \Phi(i) \end{aligned} \quad (24)$$

where $B = \Theta \otimes I_M$, $\Theta$ is weighting matrix, where $\{\Theta\}_{lk} = \delta_{lk}$, $\otimes$ denotes Kronecker product, $\Upsilon(i)$ is the diagonal matrix and $\Omega(i)$ is defined by

$$\Omega(i) = \left\{ \begin{array}{l} \dfrac{1}{\sqrt{2\pi}\sigma^2} \exp(-\dfrac{(e_1(i))^2}{2\sigma^2}) I_M, \dfrac{1}{\sqrt{2\pi}\sigma^2} \exp(-\dfrac{(e_2(i))^2}{2\sigma^2}) I_M, \cdots \\ \dfrac{1}{\sqrt{2\pi}\sigma^2} \exp(-\dfrac{(e_N(i))^2}{2\sigma^2}) I_M \end{array} \right\} \quad (25)$$

With the above set of equations, the mean and mean square analysis of the ATCDMCC algorithm can be carried out.

We first give the weight error vector for node $k$ as

$$\tilde{w}_k(i) = w_o - w_k(i) \quad (26)$$

The mean analysis considers the stability of the algorithm and derives a bound on the step size that guarantees the convergence in mean. The mean square analysis derives transient and steady-state expressions for the mean square deviation (MSD). The MSD is defined as

$$\text{MSD} = E[\|\tilde{w}_k(i)\|^2] = E[\|w_{opt} - w_k(i)\|^2] \quad (27)$$

4.1 *Mean performance*

Similar to [6-11], we define a global weight error vector as

$$\tilde{W}(i) = W_o - W(i) \quad (28)$$

Since $B W_o = W_o$, by incorporating the global weight error vector into (24), we have

$$\begin{aligned}
\tilde{W}(i) &= W_o - W(i) \\
&= W_o - B\Phi(i) \\
&= W_o - B[W(i-1) + \Upsilon(i)U^T(i)(D(i) - U(i)W(i-1))] \\
&= B\tilde{W}(i-1) - B[\Upsilon(i)U^T(i)(D(i) - U(i)W(i-1))] \\
&= B\tilde{W}(i-1) - B[\Upsilon(i)U^T(i)(U(i)W_o + V(i) - U(i)W(i-1))] \\
&= B\tilde{W}(i-1) - B[\Upsilon(i)U^T(i)(U(i)\tilde{W}(i-1) + V(i))] \\
&= B[I_{MN} - \Upsilon(i)U^T(i)U(i)]\tilde{W}(i-1) - B\Upsilon(i)U^T(i)V(i)
\end{aligned} \quad (29)$$

Here, we employ the Assumption 2 to conclude that the matrix $\Upsilon(i)$ is independent of the regressor matrix $U(i)$. Consequently, we have

$$E[\Upsilon(i)U^T(i)U(i)] \cong E[\Upsilon(i)]E[U^T(i)U(i)] \quad (30)$$

where $R_U = E[U^T(i)U(i)]$ is the auto-correlation matrix of $U(i)$. Taking the expectation on both sides of (29) gives

$$\begin{aligned}
E[\tilde{W}(i)] &= B[I_{MN} - E[\Upsilon(i)]E[U^T(i)U(i)]]E[\tilde{W}(i-1)] \\
&\quad - BE[\Upsilon(i)]E[U^T(i)]E[V(i)] \\
&= B[I_{MN} - E[\Upsilon(i)]R_U]E[\tilde{W}(i-1)] \\
&\quad - BE[\Upsilon(i)]E[U^T(i)]E[V(i)]
\end{aligned} \quad (31)$$

where, by Assumption 1, the expectation of the second term of the right hand side of (31) is zero. Then, we have

$$E[\tilde{W}(i)] = B[I_{MN} - E[\Upsilon(i)]R_U]E[\tilde{W}(i-1)] \quad (32)$$

From (32), to ensure the stability in the mean, it should hold that

$$|\lambda_{\max}(B[I_{MN} - E[\Upsilon(i)]E[U^T(i)U(i)]])| = |\lambda_{\max}(BZ)| < 1 \quad (33)$$

where $Z = [I_{MN} - E[\Upsilon(i)]E[U^T(i)U(i)]]$, and $\lambda_{\max}(.)$ denotes the maximum eigenvalue of a matrix. According to the relation $\|BZ\|_2 \leq \|B\|_2 \|Z\|_2$, we derive

$$|\lambda_{\max}(BZ)| \leq \|\Theta\|_2 |\lambda_{\max}(Z)| \quad (34)$$

Since $\|\Theta\|_2 = 1$ and for non cooperative schemes, we have $B = I_{MN}$. It follows that

$$|\lambda_{\max}(BZ)| \leq |\lambda_{\max}(Z)| \quad (35)$$

The cooperation mode can enhance the stability of the system [7]. The algorithm will therefore be stable in the mean if

$$\prod_{i=0}^{n}[I_{MN} - E[\rho_k(i)]R_{u,k}] \to 0, n \to \infty \quad (36)$$

which holds true if the mean of the step size satisfies

$$0 < E[\rho_k(i)] < \frac{2}{\lambda_{\max}(R_{u,k})} \quad (37)$$

As $\rho_k(i) = \eta_k \gamma_k(i)$, we further derive

$$0 < \eta_k < \frac{2}{\lambda_{\max}(R_{u,k})E[\gamma_k(i)]} \quad (38)$$

This condition guarantees the asymptotic unbiasedness of the ATC diffusion MCC (15). If the weight $l_1$ norm of each node is smaller than $\tau$, we have

$$\begin{aligned}
|e_k(i)| &= |d_k(i) - w_k^T(i-1)u_k(i)| \leq \|w_k(i-1)\|_1 \|u_k(i)\|_1 + |d_k(i)| \\
&\leq \tau \|u_k(i)\|_1 + |d_k(i)|
\end{aligned} \quad (39)$$

It follows easily that [30]

$$0 < \eta_k < \frac{2}{\lambda_{\max}(R_{u,k})E[G_\sigma^{MCC}(\tau\|u_k(i)\|_1 + |d_k(i)|)]}, k=1,...,N \quad (40)$$

As a result, the algorithm will be stable when the step size is within the bound of (40).

**Remark 5**: The condition of (40) is similar to those in [6,10]. The only difference is the extra term $E[G_\sigma^{MCC}(\cdot)]$, namely the expectation of the error nonlinearity introduced by MCC.

### 4.2 *Mean square performance*

Next, the mean square performance of the ATC diffusion MCC is studied. Computing the weighted norm of (29) and taking the expectations, we have

$$\begin{aligned}
&E[\|\tilde{W}(i)\|_\Sigma^2] \\
&= E[\|B[I_{MN} - \Upsilon(i)U^T(i)U(i)]\tilde{W}(i-1) - B\Upsilon(i)U^T(i)V(i)\|_\Sigma^2] \\
&= E[\|\tilde{W}(i-1)\|_{B^T \Sigma B}^2] - E[\|\tilde{W}(i-1)\|_{B^T \Sigma \Gamma(i)U(i)}^2] \\
&\quad - E[\|\tilde{W}(i-1)\|_{U(i)^T \Gamma(i)^T \Sigma B}^2] \\
&\quad + E[\|\tilde{W}(i-1)\|_{U(i)^T \Gamma(i)^T \Sigma \Gamma(i)U(i)}^2] + E[V^T(i)\Gamma(i)^T \Sigma \Gamma(i)V(i)] \\
&= E[\|\tilde{W}(i-1)\|_{\tilde{\Sigma}}^2] + E[V^T(i)\Gamma(i)^T \Sigma \Gamma(i)V(i)]
\end{aligned} \quad (41)$$

where

$$\Gamma(i) = B\Upsilon(i)U^T(i) \quad (42)$$

$$\begin{aligned}
\tilde{\Sigma} &= B^T \Sigma B - B^T \Sigma B \Upsilon(i)U^T(i)U(i) - U(i)^T U(i)\Upsilon^T(i)B^T \Sigma B \\
&\quad + U(i)^T U(i)\Upsilon^T(i)B^T \Sigma B \Upsilon(i)U^T(i)U(i) \\
&= B^T \Sigma B - B^T \Sigma \Gamma(i)U(i) - U(i)^T \Gamma(i)^T \Sigma B \\
&\quad + U(i)^T \Gamma(i)^T \Sigma \Gamma(i)U(i)
\end{aligned} \quad (43)$$

Using the data independence assumption [31] and applying the expectation operator, we get

$$\begin{aligned}
E[\tilde{\Sigma}] &= B^T \Sigma B - B^T \Sigma E[\Gamma(i)U(i)] - E[U(i)^T \Gamma(i)^T]\Sigma B \\
&\quad + E[U(i)^T \Gamma(i)^T]\Sigma E[\Gamma(i)U(i)] \\
&= B^T \Sigma B - B^T \Sigma B E[\Upsilon(i)]E[U^T(i)U(i)] \\
&\quad - E[U(i)^T U(i)]E[\Upsilon^T(i)]B^T \Sigma B \\
&\quad + E[U(i)^T \Gamma(i)^T]\Sigma \Gamma(i)U(i)]
\end{aligned} \quad (44)$$

For ease of notation, we denote $E[\tilde{\Sigma}] = \Sigma'$. Under Assumption 1, the auto-correlation matrix can be decomposed as

$$R_U = E[U^T(i)U(i)] = Q\Lambda Q^T \quad (45)$$

where $\Lambda$ is a diagonal matrix containing the eigenvalues for the entire network and $Q$ is a matrix containing the eigenvectors corresponding to these eigenvalues. Using this decomposition, we define $\bar{W}(i) = Q^T \tilde{W}(i)$, $\bar{U}(i) = U(i)Q$, $\bar{B} = Q^T B Q$, $\bar{\Sigma} = Q^T \Sigma Q$, $\bar{\Sigma}' = Q^T \Sigma' Q$, $\bar{\Upsilon} = Q^T \Upsilon(i)Q = \Upsilon(i)$, where the input regerssors are considered independent of each other at each node and the step size matrix $\Upsilon(i)$ is block diagonal. So it does not transform since $Q^T Q = I$. Then, one can rewrite (41) as

$$E[\|\tilde{W}(i)\|_{\bar{\Sigma}}^2] = E[\|\tilde{W}(i-1)\|_{\bar{\Sigma}'}^2] + E[V^T(i)\bar{\Gamma}(i)^T \Sigma \bar{\Gamma}(i)V(i)] \quad (46)$$

where

$$\begin{aligned}
\bar{\Sigma}' &= \bar{B}^T \bar{\Sigma} \bar{B} - \bar{B}^T \bar{\Sigma} \bar{B} E[\Upsilon(i)]E[\bar{U}^T(i)\bar{U}(i)] \\
&\quad - E[\bar{U}(i)^T \bar{U}(i)]E[\Upsilon^T(i)]\bar{B}^T \bar{\Sigma} \bar{B} \\
&\quad + E[\bar{U}(i)^T \bar{\Gamma}(i)^T]\Sigma \bar{\Gamma}(i)\bar{U}(i)]
\end{aligned} \quad (47)$$

where $\bar{\Gamma}(i) = \bar{B}\Upsilon(i)\bar{U}^T(i)$.

It can be seen that $E[\bar{U}^T(i)\bar{U}(i)] = \Lambda$. Using the *bvec* operator, we define $\bar{\varsigma} = bvec\{\bar{\Sigma}\}$, where $bvec\{\}$ operator divides the matrix into smaller blocks and then applies the vec operator to each of the smaller blocks. Let $R_V = \Lambda_V \odot I_M$ be the block diagonal noise covariance matrix for the entire network, where $\odot$ denotes the block Kronecker product and $\Lambda_V$ is a diagonal noise variance matrix for the network. Hence, the second term of the right hand side of (46) is

$$E[V^T(i)\bar{\Gamma}(i)^T \Sigma \bar{\Gamma}(i) V(i)] = \chi^T(i)\bar{\varsigma} \qquad (48)$$

where $\chi(i) = \text{bvec}\{R_V E[\Upsilon^2(i)]\Lambda\}$. The fourth order moment $E[\bar{U}(i)^T \bar{\Gamma}(i)^T \Sigma \bar{\Gamma}(i) \bar{U}(i)]$ in (47) remains to be evaluated. Using the step size independence assumption and the $\odot$ operator, we have

$$\text{bvec}\{E[\bar{U}(i)^T \bar{\Gamma}(i)^T]\Sigma\bar{\Gamma}(i)\bar{U}(i)]\} = (E[\Upsilon(i)\odot\Upsilon(i)]) \times A(B^T \odot B^T)\bar{\varsigma} \qquad (49)$$

According to [32], we have

$$A = diag\{A_1, A_2, \cdots, A_N\} \qquad (50)$$

in which the matrix $A_k$ is given by

$$A_k = diag\{\Lambda_1 \otimes \Lambda_k, \cdots, \lambda_k \lambda_k^T + 2\Lambda_k \otimes \Lambda_k, \cdots, \Lambda_N \otimes \Lambda_k\} \qquad (51)$$

where $\Lambda_k$ defines a diagonal eigenvalue matrix and $\lambda_k$ is the eigenvalue vector for node $k$. The output of the matrix $E[\Upsilon(i) \odot \Upsilon(i)]$ can be written as

$$\begin{aligned}
(E[\Upsilon(i) \odot \Upsilon(i)])_{kk} &= E[\text{diag}\{\rho_k(i)I_M \otimes \rho_1(i)I_M, \cdots, \\
&\quad \rho_k(i)I_M \otimes \rho_k(i)I_M, \cdots \rho_k(i)I_M \otimes \rho_N(i)I_M\}] \\
&= E[\text{diag}\{\rho_k(i)\rho_1(i)I_{M^2}, \cdots, \\
&\quad \rho^2_k(i)I_{M^2}, \cdots \rho_k(i)\rho_N(i)I_{M^2}\}] \\
&= \text{diag}\{E[\rho_k(i)]E[\rho_1(i)]I_{M^2}, \cdots, \\
&\quad E[\rho^2_k(i)]I_{M^2}, \cdots E[\rho_k(i)]E[\rho_N(i)]I_{M^2}\}]
\end{aligned} \qquad (52)$$

Now applying the *bvec* operator to the weighting matrix $\bar{\Sigma}'$ using the relation $\text{bvec}[\bar{\Sigma}'] = \bar{\varsigma}$, we can get back the original $\bar{\Sigma}'$ through $\text{bvec}[\bar{\varsigma}] = \bar{\Sigma}'$, and

$$\begin{aligned}
\text{bvec}[\bar{\Sigma}'] = \bar{\varsigma} &= [I_{M^2N^2} - (I_{MN} \odot \Lambda E[\Upsilon(i)]) - (\Lambda E[\Upsilon(i)] \odot I_{MN})] \\
&\quad + (E[\Upsilon(i) \odot \Upsilon(i)])A(B^T \odot B^T)\bar{\varsigma} = F(i)\bar{\varsigma}
\end{aligned} \qquad (53)$$

where

$$\begin{aligned}
F(i) &= [I_{M^2N^2} - (I_{MN} \odot \Lambda E[\Upsilon(i)]) - (\Lambda E[\Upsilon(i)] \odot I_{MN})] \\
&\quad + (E[\Upsilon(i) \odot \Upsilon(i)])A(B^T \odot B^T)
\end{aligned} \qquad (54)$$

Then (46) takes the following form

$$E[\|\tilde{W}(i)\|^2_{\bar{\varsigma}}] = E[\|\tilde{W}(i-1)\|^2_{F(i)\bar{\varsigma}}] + \chi^T(i)\bar{\varsigma} \qquad (55)$$

which characterizes the transient behavior of the network. Although (55) does not explicitly show the performance of the ATCDMCC, it is in fact subsumed in the weighting matrix, $F(i)$ which varies for each iteration. However, (54) clearly shows the effect of the proposed algorithm on the performance through the presence of the diagonal step size matrix $\Upsilon(i)$.

## 5. Simulation results

In order to verify the performance of the proposed DMCC algorithm in distributed network estimation case, the topology of the network with 20 nodes is generated as a realization of the random geometric graph model as shown in Figure 1. The location coordinates of the agents in the square region $[0,1.2] \times [0,1.2]$. The unknown parameter vector is set to $\frac{randn(M,1)}{\sqrt{M}}(M=10)$, where $randn(\cdot)$ is the function of generating Gaussian random. The input regressors are zero-mean Gaussian, independent in time and space with size M=10. For each simulation, the number of repetitions is set at 500 and all the results are obtained by taking the ensemble average of the network MSD over 200 independent Monte Carlo runs.

To illustrate the robust performance of the proposed algorithms, the noise at each node is assumed to be independent of the noises at other nodes, and is generated by the multiplicative model, defined as $n_k(i) = a_k(i)A_k(i)$, where $a_k(i)$ is a binary independent identically distributed occurrence process with

$p[a_k(i)=1]=c$, $p[a_k(i)=0]=1-c$, where $c$ is the arrival probability (AP); whereas $A_k(i)$ is a process uncorrelated with $a_k(i)$. The variance of $A_k(i)$ is chosen to be substantially greater (possibly infinite) than that of $a_k(i)$ to represent the impulsive noise. In this paper, we consider $A_k(i)$ as an alpha-stable noise. The alpha-stable distribution as an impulsive noise model is widely applied in the literature [14-15]. The characteristic function of alpha-stable process is defined by

$$f(t) = \exp\{j\delta t - \lambda |t|^\alpha [1+j\beta \operatorname{sgn}(t) S(t,\alpha)]\} \quad (56)$$

in which

$$S(t,\alpha) = \begin{cases} \tan\dfrac{\alpha\pi}{2} & \text{if } \alpha \neq 1 \\ \dfrac{2}{\pi}\log|t| & \text{if } \alpha = 1 \end{cases} \quad (57)$$

where $\alpha \in (0,2]$ is the characteristic factor, $-\infty < \delta < +\infty$ is the location parameter, $\beta \in [-1,1]$ is the symmetry parameter, and $\lambda > 0$ is the dispersion parameter. The characteristic factor $\alpha$ measures the tail heaviness of the distribution. The smaller $\alpha$ is, the heavier the tail is. In addition, $\lambda$ measures the dispersion of the distribution, which plays a role similar to the variance of Gaussian distribution. And then the parameters vector of the noise model is defined as $V_{\alpha-stable}(\alpha,\beta,\gamma,\delta)$.

Unless otherwise mentioned, we set the AP at 0.2, and $V_{\alpha-stable}(1.2,0,1,0)$ in the simulations below. Furthermore, we set the linear combination coefficients employing the Metropolies rule [33].

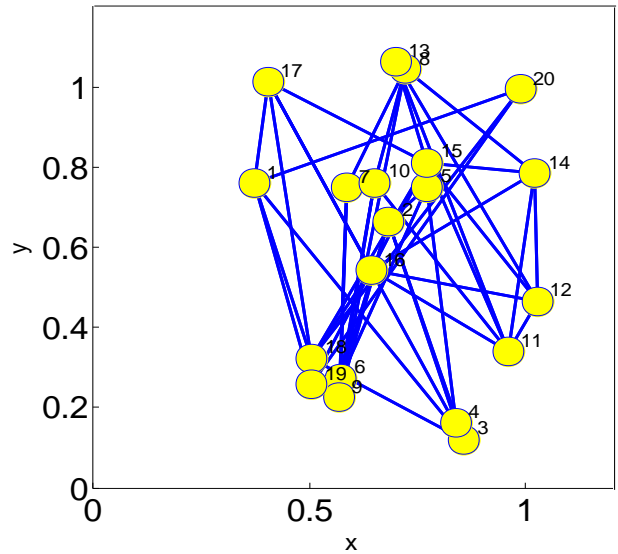

**Figure 1.** Network topology with N=20 nodes

5.1 *Performance comparison among the new methods and other algorithms*

First, the proposed algorithms (ATCDMCC and CTADMCC) are compared with some existing algorithms, including the non cooperation LMS, the ATC and CTA DLMS, the DRLS, the DLMP (including ATCDLMP and CTADLMP), and DMEE. Among these algorithms, the DLMP and DMEE algorithms can also address the DNE problem in an impulsive noise environment. To guarantee almost the same initial convergence rate, we set the step-sizes at 0.03, 0.06, 0.06 for the mentioned LMS based diffusion, DMCC and DMEE algorithms, respectively. The p is 1.2 for DLMP algorithm. Further, the kernel size is chosen as 1.0 for DMCC and DMEE algorithms. The window length is *L*=8 for DMEE. All parameters are set by scanning for the best results. Figure 2 shows the convergence curves in terms of MSD. One can observe that the convergence curve of the DLMP, DMEE and DMCC work well when large outliers occur, while other mentioned algorithms fluctuate dramatically due to the sensitivity to the impulsive noises. As can be seen from the results, the proposed DMCC algorithm has excellent performance in convergence rate and accuracy compared with other methods. The results confirm that the proposed algorithm exhibits a significant improvement in robust performance in impulsive noise environments. The steady-state MSDs at each node $k$ are shown in Figure 3. As expected, the ATC diffusion MCC algorithm performs better than all other algorithms. Although the performance of DMCC is very close to that of DMEE, its computational complexity is much lower. For this reason, we conclude that the proposed DMCC makes more sense than DMEE for applications in practice. In the subsequent simulations, we omit the results of ATCDLMS, CTADLMS, DRLS and NOCORPORATION because they often don't convergence in an impulsive noise environment.

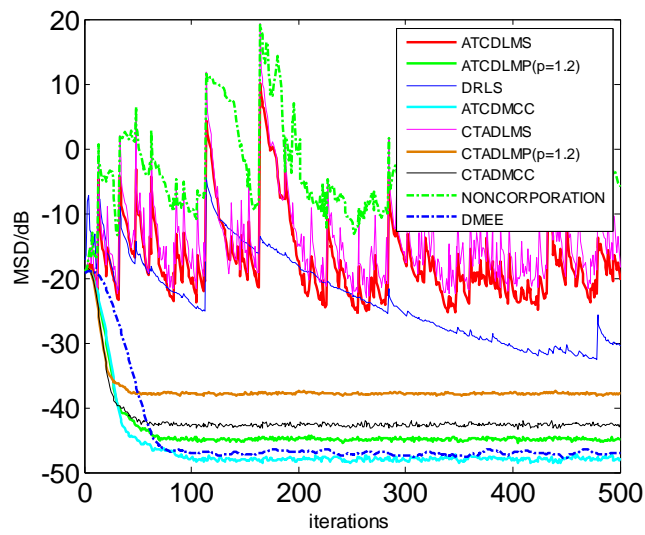

**Figure 2.** Convergence curves in terms of MSD

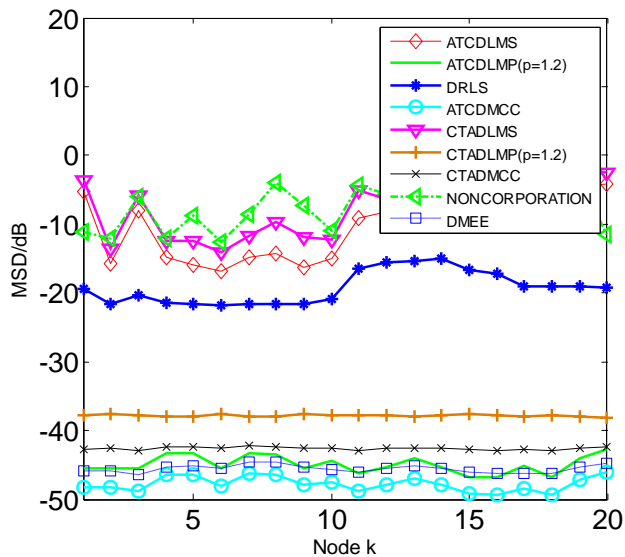

**Figure 3.** MSD at steady-state for 20 nodes

Second, we compare the performance of the proposed DMCC with that of the DLMP under different p value in terms of the MSD to show the robust performance. The p values of DLMP are selected at 1,1.1,1.2,1.4, and 2, respectively. The other parameters for the algorithms keep the same as those in the first simulation. The convergence curves in terms of MSD are shown in Figure 4. One can observe that the DLMP and DMCC work well under the impulsive noise disturbances. The results confirm the fact that the DLMP (with smaller p values) and DMCC are robust to the impulsive noises (especially with large outliers). Furthermore, the steady-state MSDs of the DLMP and DMCC algorithms are shown in Figure 5. As expected, the ATC and CTA diffusion MCC algorithms perform better than the ATC and CTA DLMP algorithms. We see that the DMCC outperforms the DLMP algorithms in that it achieves a lower steady-state MSD at each node. This result can be explained by that the MCC contains an exponential term, which reduces the influence of the large outliers significantly.

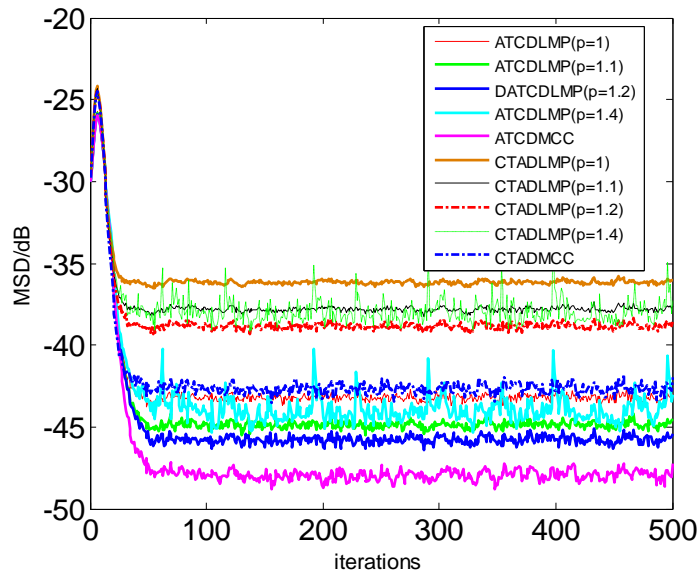

**Figure 4.** Convergence curves in terms of MSD

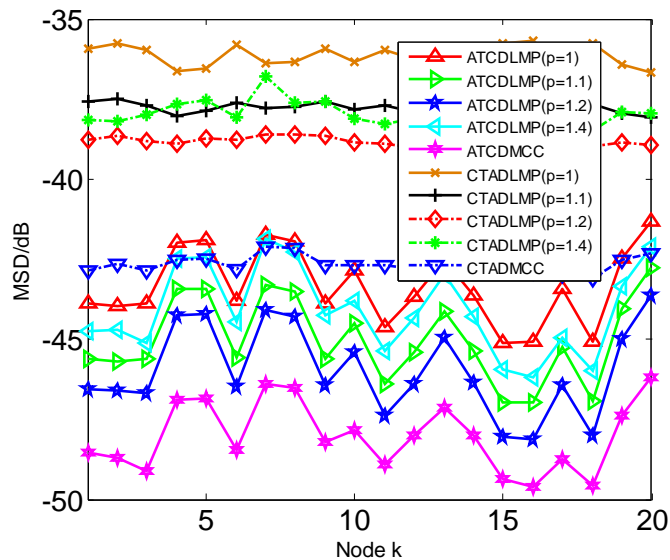

**Figure 5.** MSD at steady-state for 20 nodes

Third, we show how the exponential parameter $\alpha$ in the noise model affects the performance. From the above simulation results, we know that the ATC version diffusion algorithm is better than the CTA version. So, we compare only the performance of ATCDLMP and ATCDMCC. We set the exponential parameter $\alpha$ at 1, 1.2, 1.3, 1.4, 1.5, 1.6, 1.7 and 1.8 respectively. The other experimental settings are the same as in the previous simulation. The steady-state MSDs averaged over the last 100 iterations for different $\alpha$ values are plotted in Figure 6. It is evident that the ATCDMCC is robust consistently for different $\alpha$ values. The performance of the ATCDLMP (p=2) becomes better and better when $\alpha$ is increasing from 1.0 to 1.8. This is because that the alpha-stable distribution approaches Gaussian distribution when $\alpha$ is close to 2.0.

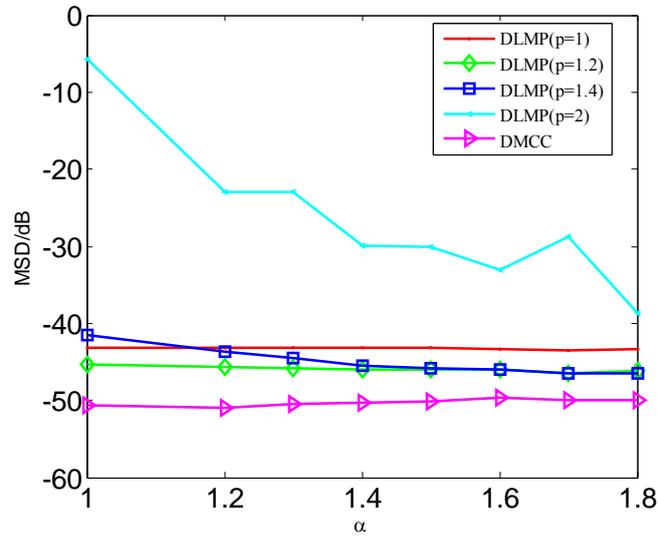

**Figure 6.** Steady-state MSD of different algorithms

Fourth, we compare the performance of the ATCMCC algorithm with the DMEE with different window lengths (5,6,8,10,12). We set M=5. For keeping the same initial convergence rate, we set the step size at 0.05 for DMEE ($L$=5,6,8,10), and 0.06 for DMEE ($L$=12) and ATCDMCC. Figure 7 shows the convergence curves of DMEE with different values of $L$ and DMCC. We observe that the ATCDMCC algorithm exhibits better performance than the DMEE ($L$=6,8,10,12), while they achieve almost the same performance when $L$=5 for DMEE. From the results we can see that the window length has important effects on the performance of DMEE (seen also detailed analysis in[17]), which will bring a hard problem of the parameter selection. Thus, the DMCC has more advantage in addressing DNE in impulsive noise environments.

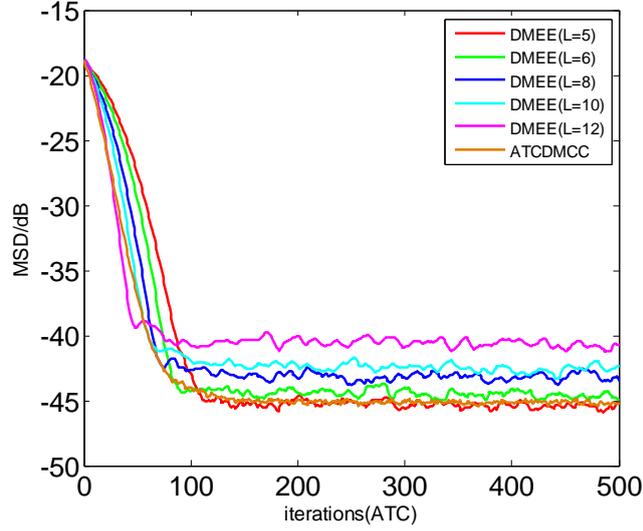

**Figure 7.** Convergence curves of ATCDMCC and DMEE with different window lengths L

5.2 *Performance of DMCC with different parameters*

First, we show how the kernel size affects the performance. The kernel size is a key parameter for the proposed diffusion version MCC algorithms (ATC and CTA DMCC). Suppose the step sizes of the proposed algorithms used at each node *k* are set at $\eta_k = 0.08$. Figure 8 shows the convergence curves of each algorithm in terms of the network MSD with different kernel sizes. One can observe that in this example, when kernel size is 1.0, both the ATC and CTA version algorithms perform very well.

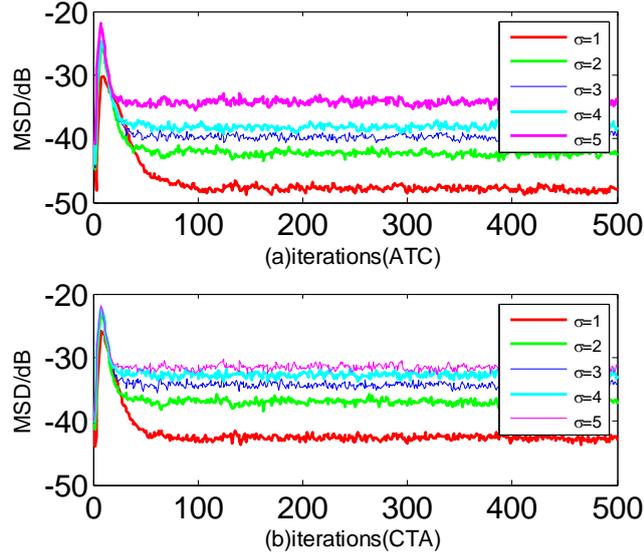

**Figure 8.** Convergence curves of DMCC DMCC with different $\sigma$

Second, we investigate how the parameter c in the noise model affects the performance of DMCC. We set the c value at 0.1, 0.2, 0.4, and 0.8, respectively. The step-size and kernel size are 0.8 and 1.0, respectively. The convergence curves with different c values are shown in Figure 9. As one can see, the steady-state MSD is increasing with the c value increasing. This is because that the outliers will occur more and more frequently when the c value becomes larger.

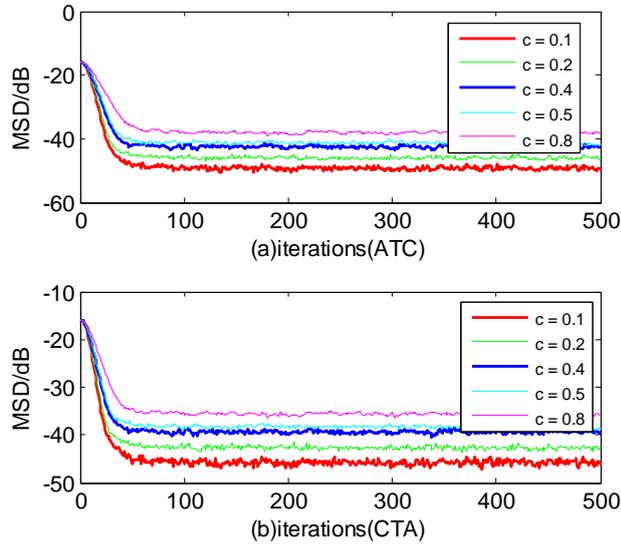

**Figure 9.** Convergence curves of DMCC with different c values

Finally, we show the joint effects of the kernel size $\sigma$ (1,2,3,4,5,6,) and noise power in terms of different $\alpha$ (1,1.2,1.4,1.6,1.8,2) on the performance. We mainly evaluate the ATC diffusion MCC algorithm in the remaining simulations. The other parameters are the same as those in the above simulations. The steady-state MSDs are shown in Figure10, from which one can see that a smaller kernel size is particularly useful for a noise with smaller $\alpha$.

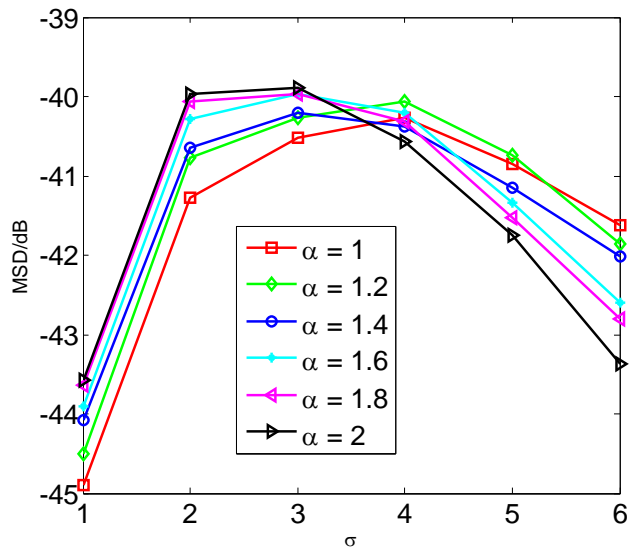

**Figure 10.** Steady-state MSD of the ATCDMCC

## 6. Conclusions

In this paper, two robust MCC based diffusion algorithms, namely the ATC and CTA diffusion MCC algorithms, are developed to improving the performance of the distributed estimation over network in impulsive noise environments. The new algorithms show strong robustness against impulsive disturbances as MCC is very effective to handle non-Gaussian noises with large outliers. Mean and mean square convergence analysis has been carried out, and a sufficient condition for ensuring the mean square stability is obtained. Simulation results illustrate that the MCC based diffusion algorithms perform very well. Especially, the ATCDMCC can achieve better performance than the robust DLMP algorithm in terms of the MSD. Although DMEE with proper $L$ can achieve almost the same performance as that of ATCMCC, its computational complexity is much higher.

## Acknowledgments

This work was supported by the 973 Program (2015CB351703) and the National Natural Science Foundation of China (No. 61372152, No. 61371807).

## References

[1] D. Estrin, G. Pottie, and M. Srivastava., Instrumenting the world with wireless sensor networks, *In Proc. IEEE Int. Conf. Acoustics, Speech, signal Processing (ICASSP)*, Salt Lake City, UT, May 2001, pp. 2033-2036.
[2] D. Li, K. D. Wong, Y. H. Hu, and A. M. Sayeed. , Detection, classification, and tracking of targets, *IEEE Signal Processing Magazine*, 19(2)(2002) 17-29.
[3] I. Akyildiz, W. Su, Y. Sankarasubramaniam, and E. Cayirci. , A survey on sensor networks, *IEEE Communications magazine*, 40(8)(2002) 102–114.
[4] L. A. Rossi, B. Krishnamachari, and C.-C. J. Kuo., Distributed parameter estimation for monitoring diffusion phenomena using physical models, *In Proc. IEEE Conf. Sensor Ad Hoc Comm. Networks*, Santa Clara, CA, Oct. 2004, pp. 460–469.
[5] F. Cattivelli and A. H. Sayed. , Diffusion LMS strategies for distributed estimation, *IEEE Transactions on Signal Processing*, 58(3)(2010)1035–1048.
[6] C. G. Lopes and A. H. Sayed. , Diffusion least-mean squares over adaptive networks: formulation and performance analysis, *IEEE Transactions on Signal Processing*, 56(7)(2008) 3122–3136.
[7] N. Takahashi, I. Yamada, A H. Sayed, Diffusion least-mean squares with adaptive combiners. IEEE International Conference on Acoustics, Speech and Signal Processing, 2009. ICASSP 2009. IEEE, 2009: 2845-2848.
[8] Takahashi N, Yamada I, Sayed A H., Diffusion least-mean squares with adaptive combiners: Formulation and performance analysis, *IEEE Transactions on Signal Processing*, 58 (9)(2010) 4795–4810.
[9] F. S. Cattivelli, C. G. Lopes, and A. H. Sayed., Diffusion recursive least-squares for distributed estimation over adaptive networks, *IEEE Transactions on Signal Processing*, 56(5)(2008)1865–1877.
[10] Saeed M O B, Zerguine A, Zummo S A., A variable step-size strategy for distributed estimation over adaptive networks, *EURASIP Journal on Advances in Signal Processing*, (1)(2013)1–14.
[11] Lee H S, Kim S E, Lee J W, et al. , A Variable Step-Size Diffusion LMS Algorithm for Distributed Estimation, *IEEE Transactions on Signal Processing*, 63(7)1808–1820.
[12] Liu Y, Li C, Zhang Z., Diffusion sparse least-mean squares over networks, *IEEE Transactions on Signal Processing*, 60(8)(2012)4480–4485.
[13] Meng-Li Cao,Qing-Hao Meng, Ming Zeng, Biao Sun, Wei Li, Cheng-Jun Ding, Distributed Least-Squares Estimation of a Remote Chemical Source via Convex Combination in Wireless Sensor Networks, *Sensors,14(* **2014**) 11444-11466.
[14] Shao, M. and Nikias, C. L., Signal processing with fractional lower order moments: stable processes and their applications, *Proceedings of the IEEE*, 81(7)(1993)986–1010.
[15] WEN F., Diffusion least-mean P-power algorithms for distributed estimation in alpha-stable noise environments, *Electronics letter*s, 49(21)(2013)1355–1356.
[16] Principe J C., "Information theoretic learning: Renyi's entropy and kernel perspectives, Springer Science & Business Media, 2010.
[17] Li C, Shen P, Liu Y, et al., Diffusion information theoretic learning for distributed estimation over network, *IEEE Transactions on Signal Processing*, 61(16)(2013)4011–4024.
[18] Weifeng Liu, Puskal P. Pokharel, and Jose C. Principe., Correntropy: Properties and Applications in Non-Gaussian Signal Processing, *IEEE Transactions on Signal Processing*, 55(11)(2007) 5286–5298.
[19] Badong Chen, José C. Príncipe., Maximum Correntropy Estimation Is a Smoothed MAP Estimation, *IEEE Signal Processing Letters*, 19(8)(2012) 491–494.
[20] Zhong F, Li D, Zhang J., Robust locality preserving projection based on maximum correntropy criterion, *Journal of Visual Communication and Image Representation*, 25(7)(2014) 1676–1685.
[21] Xing H J, Ren H R., Regularized correntropy criterion based feature extraction for novelty detection, *Neurocomputing*,133(2014) 483–490.

[22] Chen X, Yang J, Liang J, Ye Q., Recursive robust least squares support vector regression based on maximum correntropy criterion, *Neurocomputing*, 97(2012) 63–73.
[23] Abhishek Singh, Jose C. Principe., Using Correntropy as Cost Function in Adaptive Filters, *In Proceedings of International Joint Conference on Neural Networks*, Atlanta, GA , June 2009, pp. 2950–2955.
[24] Songlin Zhao, Badong Chen, Jose C. Principe , Kernel Adaptive Filtering with Maximum Correntropy Criterion, *In Proceedings of International Joint Conference on Neural Networks*, San Jose, CA, Aug 2011, pp. 2012–2017.
[25] Qu Hua, Ma Wentao, Zhao Jihong, Wang Tao, A New Learning Algorithm Based on MCC for Colored Noise Interference Cancellation, *Journal of Information & Computational Science*, 10(7)(2013)1895–1905.
[26] Ma W, Qu H, Gui G, et al., Maximum correntropy criterion based sparse adaptive filtering algorithms for robust channel estimation under non-Gaussian environments, *Journal of the Franklin Institute*, 352(7)(2015) 2708–2727.
[27] Chen B, Xing L, Liang J, Zheng N, J.C. Principe., Steady-state mean-square error analysis for adaptive filtering under the maximum correntropy criterion, *IEEE Signal Processing Letters*, 21(7)(2014) 880–884.
[28] Wu Z, Peng S, Chen B, Zhao H, Robust Hammerstein Adaptive Filtering under Maximum Correntropy Criterion. *Entropy,* 2015, 17(10), 7149-7166.
[29] Lee J W, Kim S E, Song W J., Data-selective diffusion LMS for reducing communication overhead, *Signal Processing*, 113(2015)211-217.
[30] Chen B, Wang J, Zhao H, et al., Convergence of a Fixed-Point Algorithm under Maximum Correntropy Criterion, *IEEE Signal Processing Letters*, 22(10)(2015)1723–1727.
[31] AH Sayed., Fundamentals of Adaptive Filtering, Wiley, New York, 2003.
[32] AI Sulyman, A Zerguine., Convergence and steady-state analysis of a variable step-size NLMS algorithm, *Signal Processing*, 83(6)(2003)1255–1273.
[33] Xiao L, Boyd S., Fast linear iterations for distributed averaging, *Systems & Control Letters*, 53(1)(2004)65–78.